\documentclass[conference]{IEEEtran}
\IEEEoverridecommandlockouts
\usepackage{cite}
\usepackage{amsmath,amssymb,amsfonts}
\usepackage{algorithmic}
\usepackage{graphicx}
\usepackage{textcomp}
\usepackage{enumitem}

\usepackage{xcolor}
\def\BibTeX{{\rm B\kern-.05em{\sc i\kern-.025em b}\kern-.08em
    T\kern-.1667em\lower.7ex\hbox{E}\kern-.125emX}}
\begin{document}

\title{LLMSense: Harnessing LLMs for High-level Reasoning Over Spatiotemporal Sensor Traces
}

\author{\IEEEauthorblockN{Xiaomin Ouyang}
\IEEEauthorblockA{
University of California, Los Angeles\\
xmouyang23@ucla.edu}
\and
\IEEEauthorblockN{Mani Srivastava}
\IEEEauthorblockA{
University of California, Los Angeles and Amazon$^{*}$\thanks{$^{*}$Mani Srivastava holds concurrent appointments as a Professor of ECE and CS (joint) at UCLA and as an Amazon Scholar. This paper describes work performed at UCLA and is not associated with Amazon.}\\
mbs@ucla.edu}
}


\maketitle

\pagenumbering{arabic} 
\thispagestyle{plain}
\pagestyle{plain}

\begin{abstract}
\noindent
Most studies on machine learning in sensing systems focus on \emph{low-level perception tasks} that process raw sensory data within a short time window. However, many practical applications, such as human routine modeling and occupancy tracking, require \emph{high-level reasoning} abilities to comprehend concepts and make inferences based on long-term sensor traces. Existing machine learning-based approaches for handling such complex tasks struggle to generalize due to the limited training samples and the high dimensionality of sensor traces, necessitating the integration of human knowledge for designing first-principle models or logic reasoning methods. We pose a fundamental question: Can we harness the reasoning capabilities and world knowledge of Large Language Models (LLMs) to recognize complex events from long-term spatiotemporal sensor traces? To answer this question, we design an effective prompting framework for LLMs on high-level reasoning tasks, which can handle traces from the raw sensor data as well as the low-level perception results. We also design two strategies to enhance performance with long sensor traces, including summarization before reasoning and selective inclusion of historical traces. Our framework can be implemented in an edge-cloud setup, running small LLMs on the edge for data summarization and performing high-level reasoning on the cloud for privacy preservation. The results show that LLMSense can achieve over 80\% accuracy on two high-level reasoning tasks such as dementia diagnosis with behavior traces and occupancy tracking with environmental sensor traces. This paper provides a few insights and guidelines for leveraging LLM for high-level reasoning on sensor traces and highlights several directions for future work.
\end{abstract}


\section{Introduction}
\label{sec:introduction}

The wealth of sensor data collected by mobile and Internet of Things (IoT) devices has the representation of the physical world. These IoT sensors, when integrating with machine learning techniques, could create an environment with ambient intelligence. Most of the current studies on machine learning in sensing systems are focused on \emph{low-level perception tasks} that process raw sensory data a relatively short time window to make predictions about the current state. For example, extracting features or data patterns from 2-second sensor data for activity recognition, object detection, and speech recognition. These tasks are more concerned with sensory input and basic recognition rather than higher-level interpretation.

\begin{figure}
    \setlength{\abovecaptionskip}{0.cm}
    \centering
     \includegraphics[width = \linewidth]{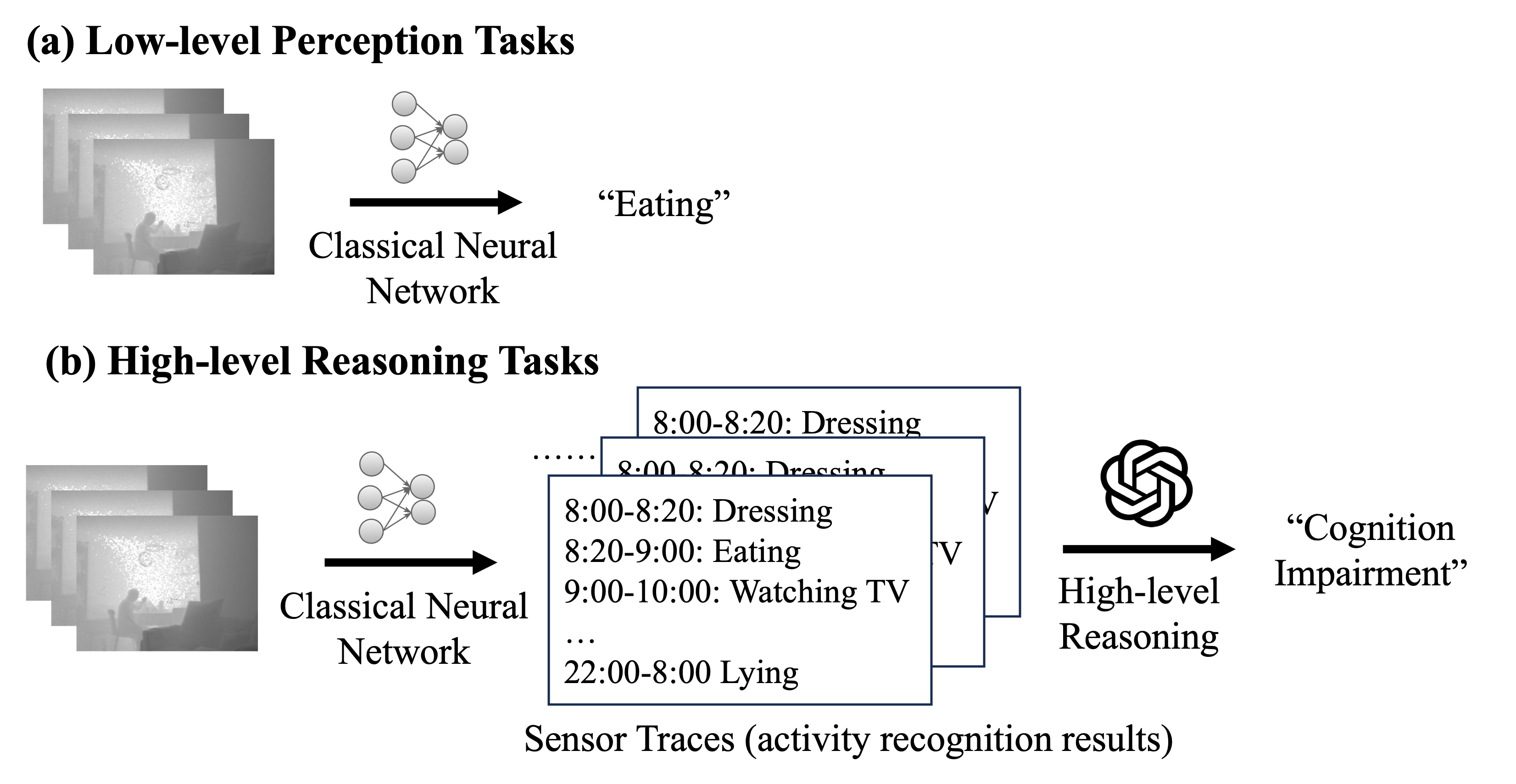}
  \caption {Comparison between low-level perception tasks (e.g., activity recognition) and high-level reasoning tasks (cognitive impairment detection).}
  \label{fig:intro-figure}
  \vspace{-10pt}
\end{figure}

However, many practical applications also require understanding concepts and making inferences based on long-term sensor data, which we named \emph{high-level reasoning tasks}. For example, as shown in Figure \ref{fig:intro-figure}, based on the longitudinal perception results of activity recognition, we could induce whether the subject has cognitive impairment or not. Other high-level reasoning tasks based on sensor traces include occupancy tracking \cite{weber2023overcoming}, human routine modeling \cite{banovic2016modeling, sun2014nonparametric}, environmental monitoring \cite{ingelrest2010sensorscope}, and smart energy management \cite{han2020efficient}. Such tasks typically require sophisticated reasoning ability to interpret complex sensor traces and integrate domain knowledge to make predictions or decisions. Most of the existing approaches on high-level reasoning tasks are either based on training machine learning models with the input of long-term sensor traces or applying first-principle rules to aggregate low-level perception results. However, studies show that machine learning-based approaches do not generalize well on data collected from different environments \cite{weber2023overcoming} or populations \cite{xu2023globem}, , due to the limited training samples and the high dimensionality of sensor traces. Designing first-principle models for aggregating the short-time perception results require carefully integrate human knowledge.

Existing studies have demonstrated the potential of large language models (LLMs) to interpret complex patterns and trends within sequential data, such as for time-series data analysis \cite{chang2023llm4ts} and as general pattern learners \cite{mirchandani2023large}. Moreover, LLMs possess a vast repository of the world and expert knowledge to perform various complex tasks, such as activity recognition \cite{liu2023large,xu2023penetrative}, root cause analysis for cloud instants \cite{chen2024automatic}, and even medical diagnosis \cite{yan2023multimodal}. 
Therefore, we pose a fundamental question:  Can we harness the reasoning capabilities and world knowledge of Large Language Models (LLMs) to analyze observations derived from long-term sensor traces for high-level reasoning? 
There are several major challenges when applying LLMs to analyze long-term sensor traces for high-level reasoning. First, how to convert and format the sensor traces into natural language sentences to be interpreted by LLMs. Second, how can we improve the performance of LLMs when handled with the long series of sensor traces. 




In this paper, we propose LLMSense, a LLM-based system for high-level reasoning on sensor traces. The core of our approach is to leverage the reasoning ability and domain knowledge of LLMs to interpret the long-term sensor traces for high-level reasoning. 
To achieve this, we propose an effective prompting framework for high-level reasoning tasks on sensor traces, which can deal with traces from the raw sensor data as well as the low-level perception results. We carefully designed the prompts to incorporate the instructions of the problem, context information (e.g., data collection, physical meaning of sensor data, and demographic information of subjects), textualized and structured sensor traces, and constraints of output formats. Moreover, we design two approaches to further enhance the performance with long sensor traces, including summarization before reasoning and selective inclusion of historical traces. Our framework can be implemented in an edge-cloud framework by running small LLMs on edge to extract summarization of traces, and perform high-level reasoning on cloud to preserve data privacy.

We evaluate the design of LLMSense on two high-level reasoning tasks on sensor traces, including dementia diagnosis with behavior traces and occupancy tracking with environmental sensor traces. The results show that LLMSense can achieve about 80\% accuracy on complex high-level reasoning tasks. Moreover, we show that LLMSense can be implemented in an edge-cloud framework to achieve a trade-off between accuracy, latency and data privacy. 

In summary, we make the following key contributions:
\begin{itemize}
    \item We present a study on leveraging language models for high-level reasoning tasks on sensor traces. We design a pipeline that converts raw sensor traces or low-level perception results into sentences for zero-shot high-level reasoning tasks.
    \item We propose LLMSense, an effective prompting framework for high-level reasoning tasks on sensor traces, and design two approaches to further enhance the performance with long sensor traces, including summarization before reasoning and selective inclusion of historical traces.
    \item We evaluate the design of LLMSense on two high-level reasoning tasks over long-term sensor traces, including dementia diagnosis and occupancy tracking. The results show that LLMSense achieves reasonably good and consistent results.
    \item We provide a few insights and guidelines for leveraging LLM for high-level reasoning on sensor traces, and highlight several directions for future work.
\end{itemize}

\section{Related Work}

\textbf{High-level reasoning on sensor traces.} Many practical applications require high-level reasoning abilities to understand concepts and making inferences based on long-term sensor data. Related tasks include occupancy tracking \cite{weber2023overcoming}, human routine modeling \cite{banovic2016modeling}, environmental monitoring \cite{ingelrest2010sensorscope}, and energy management \cite{han2020efficient}. Such tasks typically require sophisticated reasoning ability to interpret complex sensor traces and integrate domain knowledge to make prediction or decision. Most of existing approaches are either based on training machine learning models with the input of long-term sensor traces \cite{weber2023overcoming} or applying first-principle rules to aggregate low-level perception results \cite{jeyakumar2023x}. However, studies show that these approaches does not generalize well on data collected from different environments \cite{weber2023overcoming} or populations \cite{xu2023globem}.

\textbf{LLMs for time-series data analysis.} Recently, LLMs have been applied in time-series data analysis due to the close structure between time-series and text data. For example, \cite{chang2023llm4ts, jin2023time, xue2023utilizing} leverage the LLMs for time-series forecasting. \cite{mirchandani2023large} applies pre-trained LLMs on temporal sequence tasks such as modeling and extrapolation, and showed that LLMs can serve as general pattern learners. These studies showcase LLMs' ability to learn and extract complex patterns from time-series data. However, there are fewer works on the effective execution of high-level reasoning on sensor traces, which requires not only understanding the patterns of traces, but also employing substantial domain knowledge relevant to the application to interpret the sensor traces.

\textbf{LLMs for domain-specific applications.} Being trained on large amounts of language data, LLMs possess a vast repository of world knowledge and can be adapted in various domains through zero-shot learning or fine-tuning. For example, with the input of language data, LLM can be leveraged to perform various complex tasks, such as activity recognition \cite{liu2023large}, root cause analysis for cloud instants \cite{chen2024automatic}, and even medical diagnosis \cite{yan2023multimodal,xu2023leveraging}. However, how to interpret the long-term sensor traces with world knowledge in LLMs still remains an open question.


\section{Motivation}

\subsection{Application Scenarios}
\label{sec:source_traces}

We propose to leverage the pre-trained LLMs for high-level reasoning over spatial-temporal sensor traces. Our framework can be applied to two primary formats of data input: directly processing streaming sensor data or utilizing low-level perception results.

\begin{figure*}
    \setlength{\abovecaptionskip}{0.cm}
    \centering
     \includegraphics[width = \linewidth]{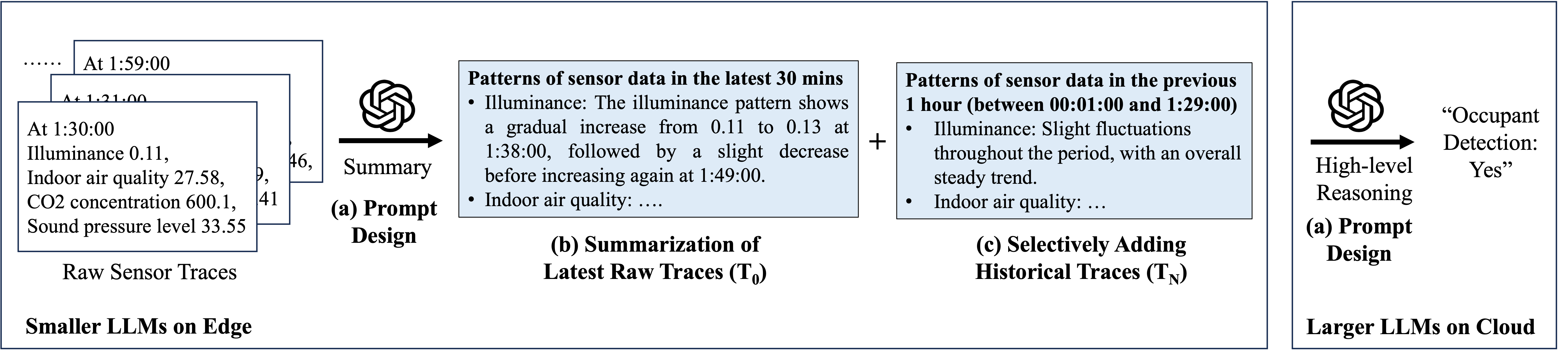}
  \caption {Overview of LLMSense. (a) Effective prompting for high-level reasoning over sensor traces. We design two approaches to enhance the performance with long sensor traces: (b) Summarization of the traces before reasoning. (c) Selective inclusion of historical traces. LLMSense can be implemented in an edge-cloud framework by running small LLMs on edge to extract summarization of traces, and perform high-level reasoning on cloud to preserve data privacy.}
  \vspace{-10pt}
  \label{fig:summary_history_reasoning}
\end{figure*}

\begin{figure}
    \setlength{\abovecaptionskip}{-0.2cm}
    \centering
     \includegraphics[width = \linewidth]{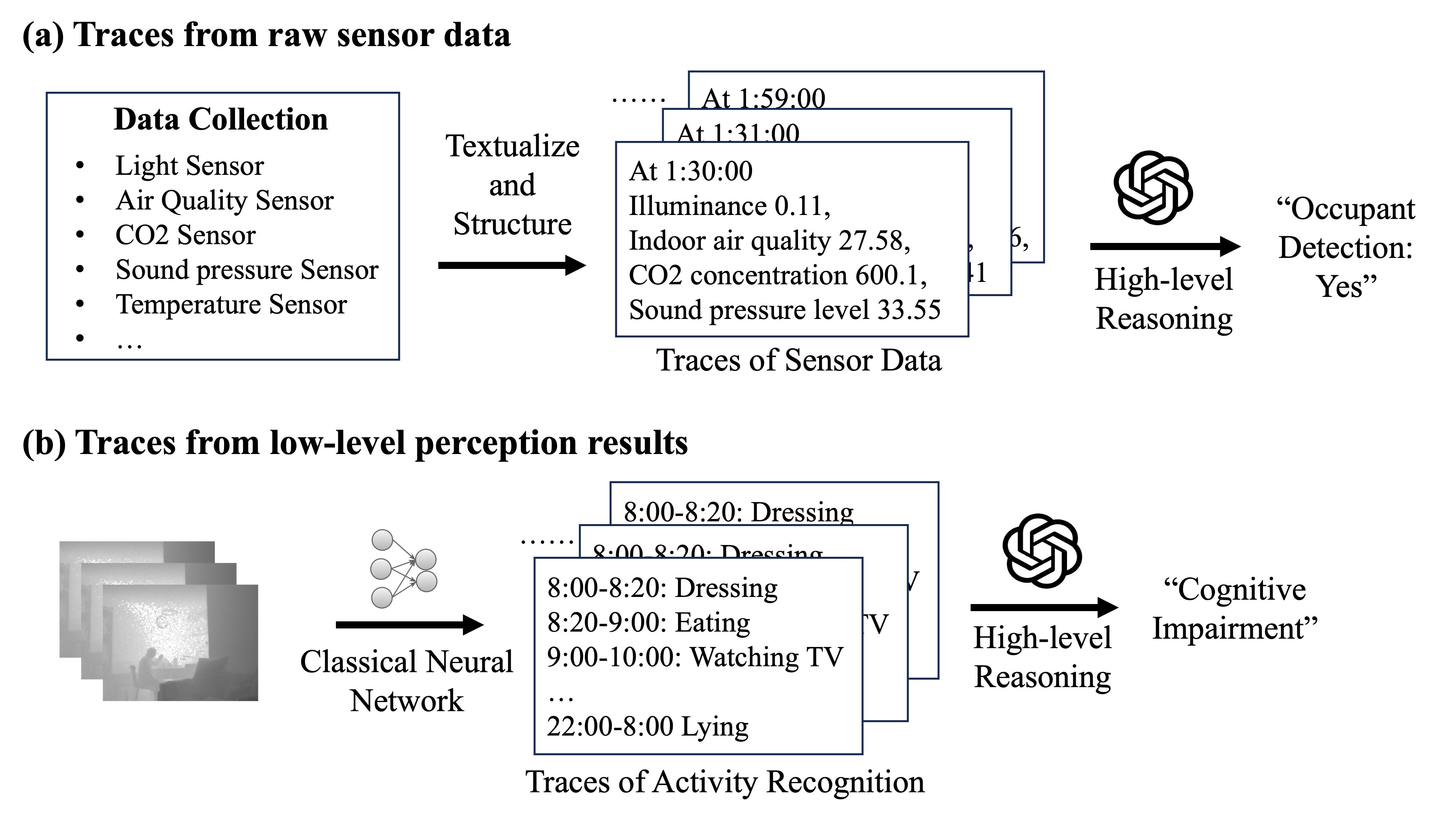}
  \caption {Different sources of spatial-temporal traces. (a) Traces from raw sensor data. (b) Traces from low-level perception results. }
  \label{fig:input_traces}
  \vspace{-10pt}
\end{figure}

\textbf{Directly processing streaming sensor data.} When the sensor data has a low data rate and it is not intuitive to define an intermediate low-level perception task, we could directly feed and transform the streaming sensor data into language models. For example, as shown in Figure \ref{fig:input_traces}, when performing occupancy tracking with environmental sensor data, we could directly textualize and structure the sensor data and input it to LLMs. In this case, language models can interpret these data streams to perform context-aware reasoning and generate predictions based on the input sensor readings.

\textbf{Utilizing low-level perception results.} When the sensor data has a high data rate such as depth videos, it will be inefficient to directly input the sensor data into LLMs. Therefore, we could first use a classical machine learning model to do low-level perception tasks, and then structure the perception results to be input to large language models. For example, as shown in Figure \ref{fig:input_traces}, in the task of cognitive impairment detection with longitudinal sensor data, we could input the traces of activity recognition results to LLMs for analyzing the behaviors of subjects.


\subsection{Advantages of leveraging LLMs}

The motivations for leveraging LLMs for high-level reasoning on sensor traces are as follows.

\textbf{Making use of world knowledge in LLMs.} LLMs possess a vast repository of the world and expert knowledge to perform various complex tasks, such as activity recognition \cite{liu2023large}, root cause analysis \cite{chen2024automatic}, and even medical diagnosis \cite{yan2023multimodal}. By integrating world knowledge, language models can better understand the physical world from sensor traces for high-level reasoning tasks. For example, cognitive impairment detection based on behavior traces of subjects requires lots of disease knowledge about the symptoms of dementia.

\textbf{The ability to understand long-series texts.} LLMs have been successfully applied as general pattern learners \cite{mirchandani2023large} and for time-series data analysis \cite{chang2023llm4ts}, showcasing their ability to understand complex patterns within sequential data. The sensor traces, either from raw sensor data or the low-level prediction results, consist of continuous measurements or observations of the physical world. By transforming the sensor traces into texts, we can harness the ability of LLMs to understand long-term sensor traces for high-level reasoning tasks.

\textbf{Generalizable ability with little training data.} Existing studies show that conventional machine learning-based approaches do not generalize well on data collected from different environments \cite{weber2023overcoming} or populations \cite{xu2023globem}. The inherent reasoning ability of LLMs enables them to extrapolate from previous knowledge and generalize to new cases, enhancing their adaptability to diverse settings.

\section{Design of LLMSense}

Figure \ref{fig:summary_history_reasoning} shows an overview of LLMSense. We design an effective prompting framework for high-level reasoning over sensor traces. Moreover we design two approaches to enhance the performance with long sensor traces: including summarization of the traces before reasoning and selective inclusion of historical traces. Our framework can be implemented in an edge-cloud framework by running small LLMs on edge to extract summarization of traces, and perform high-level reasoning on cloud to preserve data privacy.

\label{sec:approach}
\subsection{Prompt Design for High-level Reasoning on Sensor Traces}

\begin{figure}
    \setlength{\abovecaptionskip}{0.cm}
    \centering
     \includegraphics[width = \linewidth]{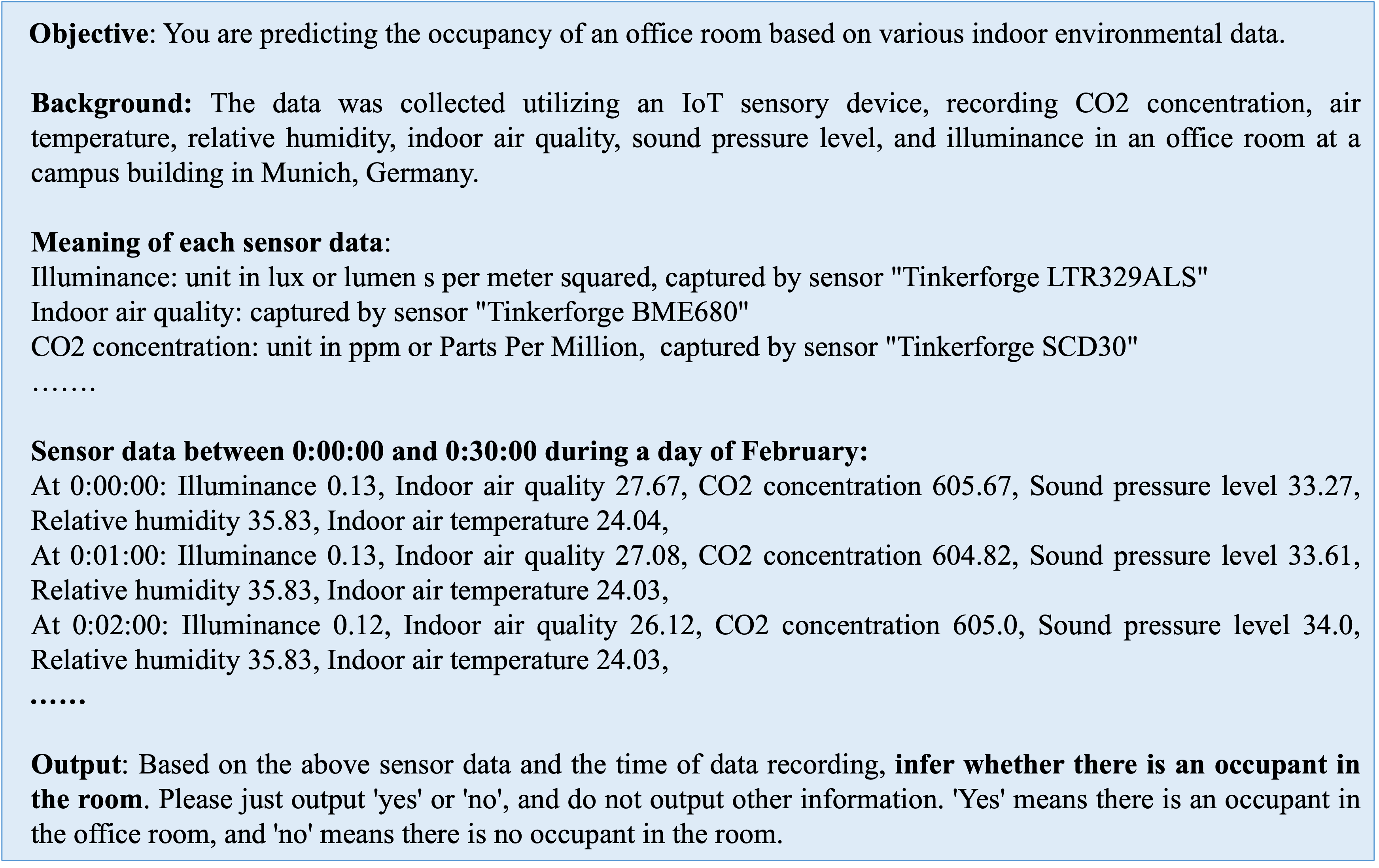}
  \caption {The prompt of occupancy tracking based on traces of environmental data in the latest 30 minutes.}
  \label{fig:basic_prompt}
  \vspace{-10pt}
\end{figure}

To improve the reasoning capability of LLMs, we carefully design a comprehensive zero-shot prompting for high-level reasoning tasks on sensor traces. The goal of prompt design is to empower a pre-trained general-purpose LLM to achieve good performance on tasks in a specific high-level reasoning task on sensor traces. We propose a general zero-shot prompt template that consists of four parts:
\begin{equation}
    Prompt = Objective + Context + Data + Format
\end{equation}
where \emph{Objective} provides the instructions for the task that LLMs are required to perform. \emph{Context} includes the background of the task and sensor traces, which inject general domain knowledge of the task and are expected to remain for different inputs of sensor traces. For example, as shown in Figure \ref{fig:basic_prompt}, to enhance the performance of the occupancy tracking task with sensor traces, we incorporate the background information of data collection, as well as the physical meaning of sensor data as context information in the prompt. \emph{Data} presents the spatial-temporal data traces, which could be extracted from the raw sensor data or the low-level perception results (see Section \ref{sec:source_traces}). \emph{Format} poses the specific questions that LLMs need to answer and constrain the output of models.


Moreover, we iteratively optimize the prompts so that the LLM models could achieve good performance on the accuracy, consistency, and uncertainty (see Section \ref{sec:metric}) of the reasoning task on several samples. Then during inference, we apply the optimized prompt template directly to a large amount of unseen samples to evaluate the generalization ability of LLMs.

\subsection{Dealing with Long Traces}

When the length of sensor traces increases, the performance of LLMs will suffer degradation due to the context limit. We propose two approaches to improve the performance of LLMs for high-level reasoning on long traces, including summarization of the traces before reasoning and selective inclusion of historical traces.

\subsubsection{Summarization before Reasoning}

When leveraging LLMs for analyzing the sensor traces, one major challenge is that the length of the sensor traces used for high-level reasoning is often too extensive. For example, to diagnose whether an elderly subject has cognitive impairment or not, we need observations of behavior symptoms for at least two or four weeks, which can take more than 5,000 tokens when transformed into language texts. The considerable number of tokens in the sensor traces can pose challenges for the LLM to effectively process the information and could introduce large uncertainty in reasoning tasks. Therefore, feeding the spatial-temporal sensor traces directly into the LLM to make a prediction could not be an ideal choice, especially when the length of the sensor traces is extremely long. 

\begin{figure}
    \setlength{\abovecaptionskip}{0.cm}
    \centering
     \includegraphics[width = \linewidth]{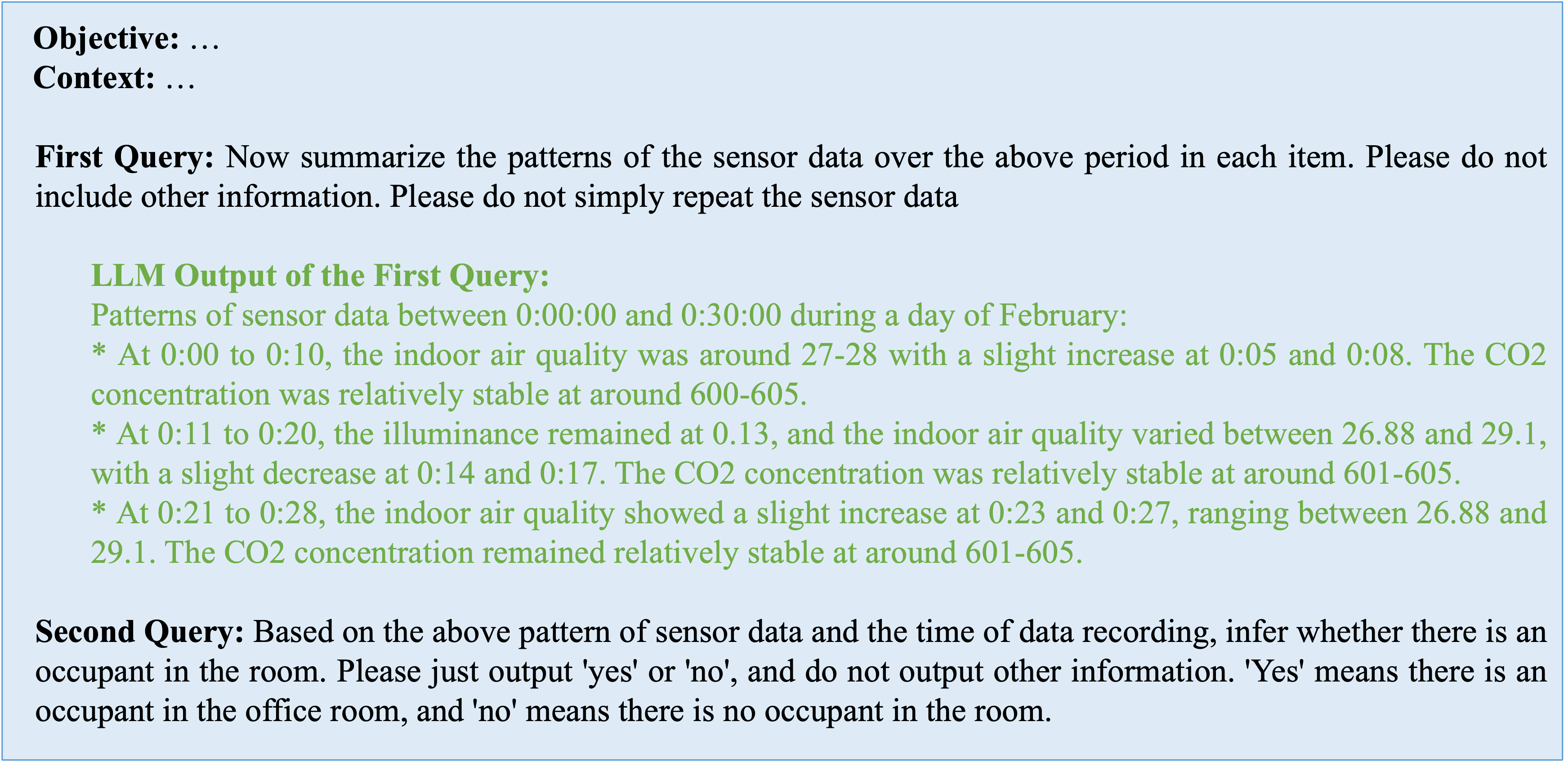}
  \caption {Example of prompts for summarization of the traces before reasoning.}
  \label{fig:summary_prompt}
  \vspace{-5pt}
\end{figure}

Therefore, we propose to leverage the summarization ability of LLMs to summarize the patterns of the sensor traces first before making high-level reasoning. For example, as shown in Figure \ref{fig:summary_prompt}, we perform two queries in the prompts, including: the first query of summarizing the patterns of the sensor traces over the given period, and the second query of occupancy tracking based on the pattern of sensor traces. The summarization process will make the information of the long sensor traces more concise and informative, and can insert domain knowledge that is useful for reasoning. Moreover, such intermediate results given by summarization could make the results more explainable.

\subsubsection{Selective Inclusion of History Traces}

\begin{figure}
    \setlength{\abovecaptionskip}{0.cm}
    \centering
     \includegraphics[width = \linewidth]{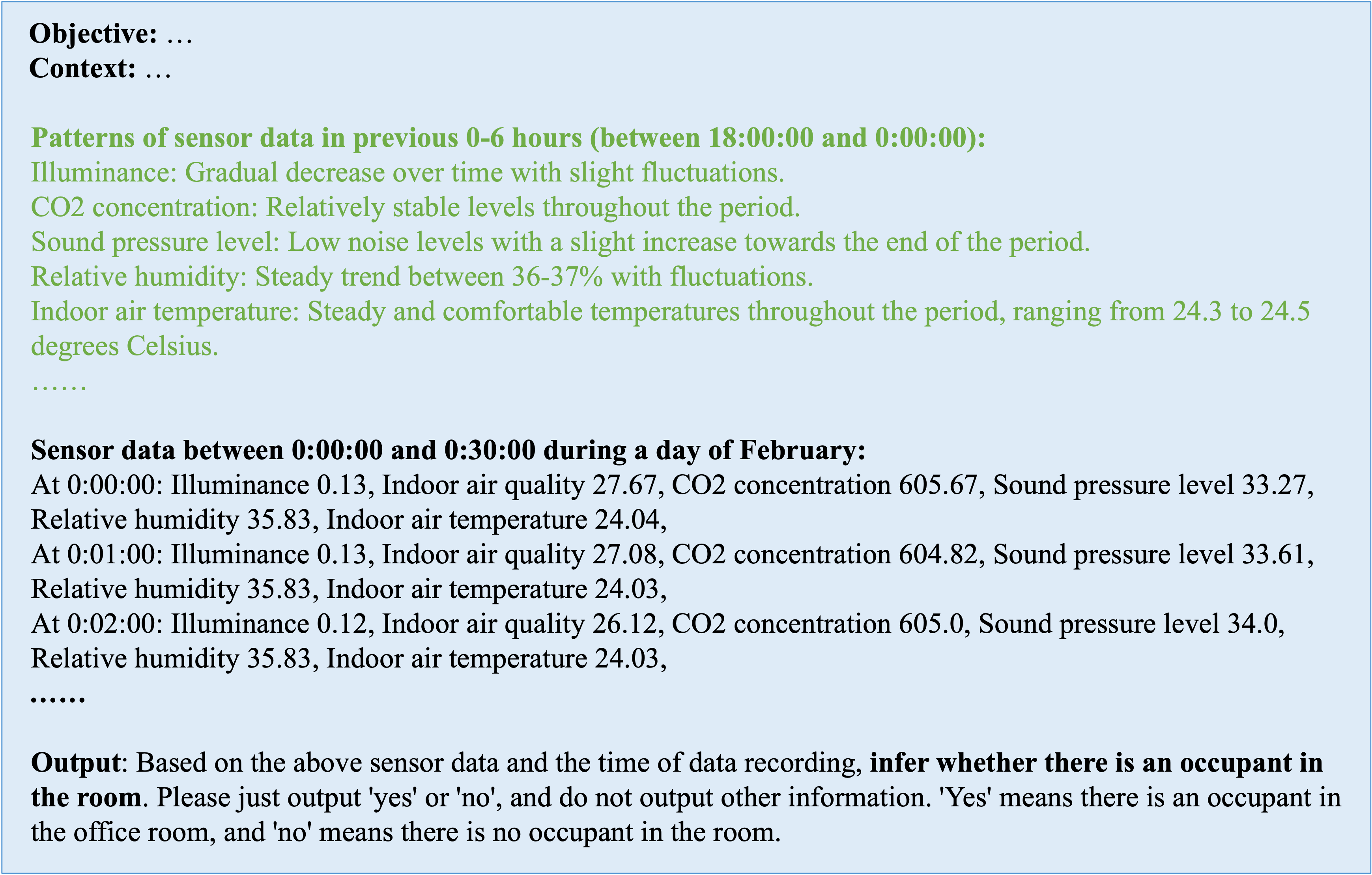}
  \caption {Example of prompts for selectively adding history traces.}
  \label{fig:history_prompt}
  \vspace{-10pt}
\end{figure}

When dealing with the extensive length of sensor traces, the summarization of LLMs may be also noisy. Therefore, we propose to selectively add historical traces to strategically incorporate only the most relevant and significant historical data points. 
This process helps to condense the information while still preserving the essential context. By prioritizing key events or patterns that provide crucial insights or context for the sensor data, analysts can streamline the analysis and focus on the most impactful historical traces. The total duration of the sensor traces is denoted as:

\begin{equation}
    Duration \, T = Latest \, T_0 + History \, T_N
\end{equation}
where $Latest \, T_0$ means traces collected in the latest $T_0$, and $History \, T_N$ means to add the information from data collected in previous $T_N$. Here $T_0$ and $T_N$ need to be assigned for different applications. For example, as shown in Figure \ref{fig:history_prompt}, we set $T_0 = 30 min$ to input the raw sensor traces from the latest 30 minutes and $T_N=6 \, hours$ incorporate the data pattern of the previous 6 hours.

\section{Evaluation and Results}
\label{sec:evaluation}

\subsection{Experiment Settings}
\subsubsection{Datasets and Tasks} We select two datasets for high-level reasoning tasks, including dementia diagnosis with behavior traces and occupancy tracking with sensor traces.

\textbf{Dementia diagnosis with behavior traces \cite{ouyang2023harmony}.} This dataset contains multimodal sensor data to detect 11 daily activities related to Alzheimer’s Disease (AD) in four continuous weeks. The data is downsampled and labeled manually. There are a total of 16 elder subjects, including six with AD, six with mild cognitive impairment (MCI), and four cognitively normal subjects. Since the data rate of sensors is very high, we use the labels and timestamps in this paper to diagnose whether the subjects have cognition impairment or not.


\textbf{Occupancy tracking with environmental sensor traces \cite{weber2023overcoming}.} This dataset contains sensor data from 80 working days in an office room of a university building in Munich, Germany. The sensor data include ambient light, sound pressure level, air temperature, indoor air quality, relative humidity, and CO2. The occupancy ground truth is documented by manual recording.  We use the sensor data and timestamps in this paper to detect whether there is occupancy in the room or not.

\subsubsection{Implementation} We evaluate the performance of the two tasks on LLama2-13B, LLama2-70B, and GPT3.5 through APIs in a zero-shot prompting. For each data sample, we run LLMs with the same prompt for five times. We also run LLama2 with ollama \cite{ollama} on our local machine (Intel Core i7-6950X CPU @ 3.00GHz × 20, 128GB) to measure the latency of on-device inference.


\subsubsection{Metrics} We evaluate the following metrics of the results predicted by LLMs, including accuracy, consistency, and uncertainty. Here accuracy measures whether the LLMs generate correct prediction results. The consistency measures whether LLMs make consistent predictions across the five trials. The results are consistent if the number of trials having the same result is no less than four. The uncertainty is defined when LLM cannot give a definite decision based on the input.
\label{sec:metric}

\subsection{Results}

\begin{figure}
    \setlength{\abovecaptionskip}{0.cm}
    \centering
     \includegraphics[width = \linewidth]{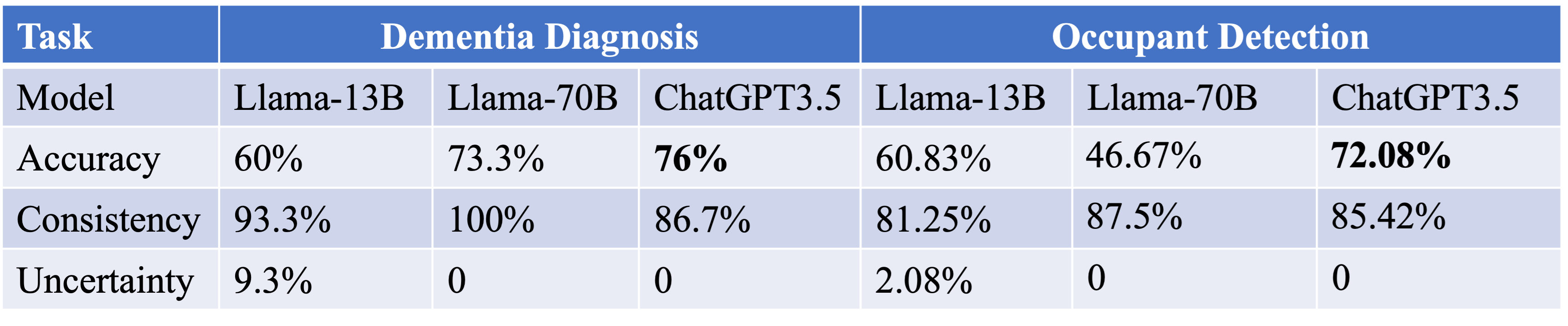}
  \caption {Overall performance of two high-level reasoning tasks with different language models.}
  \label{fig:accuracy_overall}
  \vspace{-5pt}
\end{figure}

\begin{figure}
    \setlength{\abovecaptionskip}{0.cm}
    \centering
     \includegraphics[width = \linewidth]{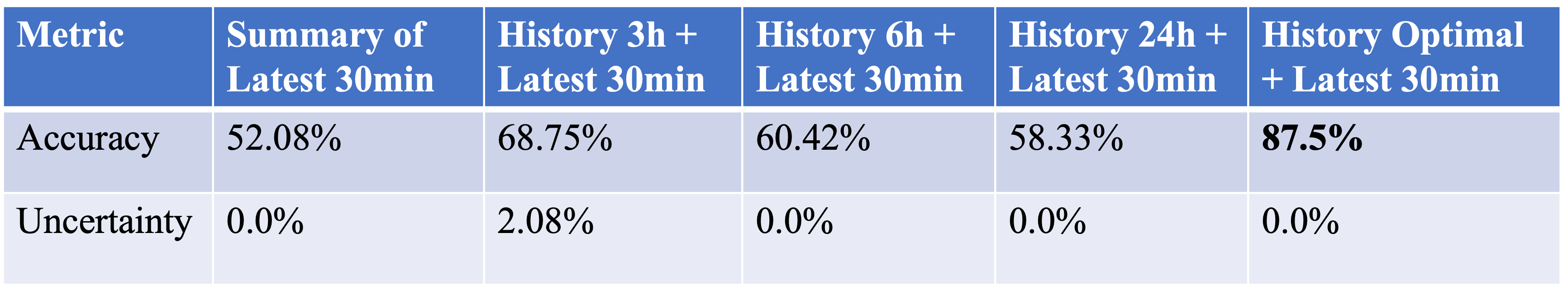}
  \caption {Performance with different lengths of historical traces.}
  \label{fig:accuracy_selective}
  \vspace{-10pt}
\end{figure}

\subsubsection{Overall Performance} Figure \ref{fig:accuracy_overall}
shows the performance of two high-level reasoning tasks with different language models. First, in zero-shot settings, GPT3.5 achieves 76\% and 72.08\% accuracy on cognitive impairment detection and occupancy tracking, which is comparable with the performance reported in previous studies that train machine learning models with lots of training data \cite{ouyang2023harmony,weber2023overcoming}. This shows that our design could leverage the reasoning ability and disease knowledge of LLMs for complex tasks on long-term sensor traces. Second, there is a trade-off between consistency and accuracy performance when using different LLMs. For example, in dementia diagnosis, the smaller models (Llama-13B and Llama-70B) have more consistent predictions, and the prediction results are less accurate than GPT3.5. Therefore, we could further develop adaptive voting mechanisms to filter inconsistent predictions of LLMs.

\begin{figure}
    \centering
    \begin{minipage}{.46\linewidth}
    \setlength{\abovecaptionskip}{0.cm}
    \setlength{\belowcaptionskip}{-0.cm}
    \centering
     \includegraphics[width = \textwidth]{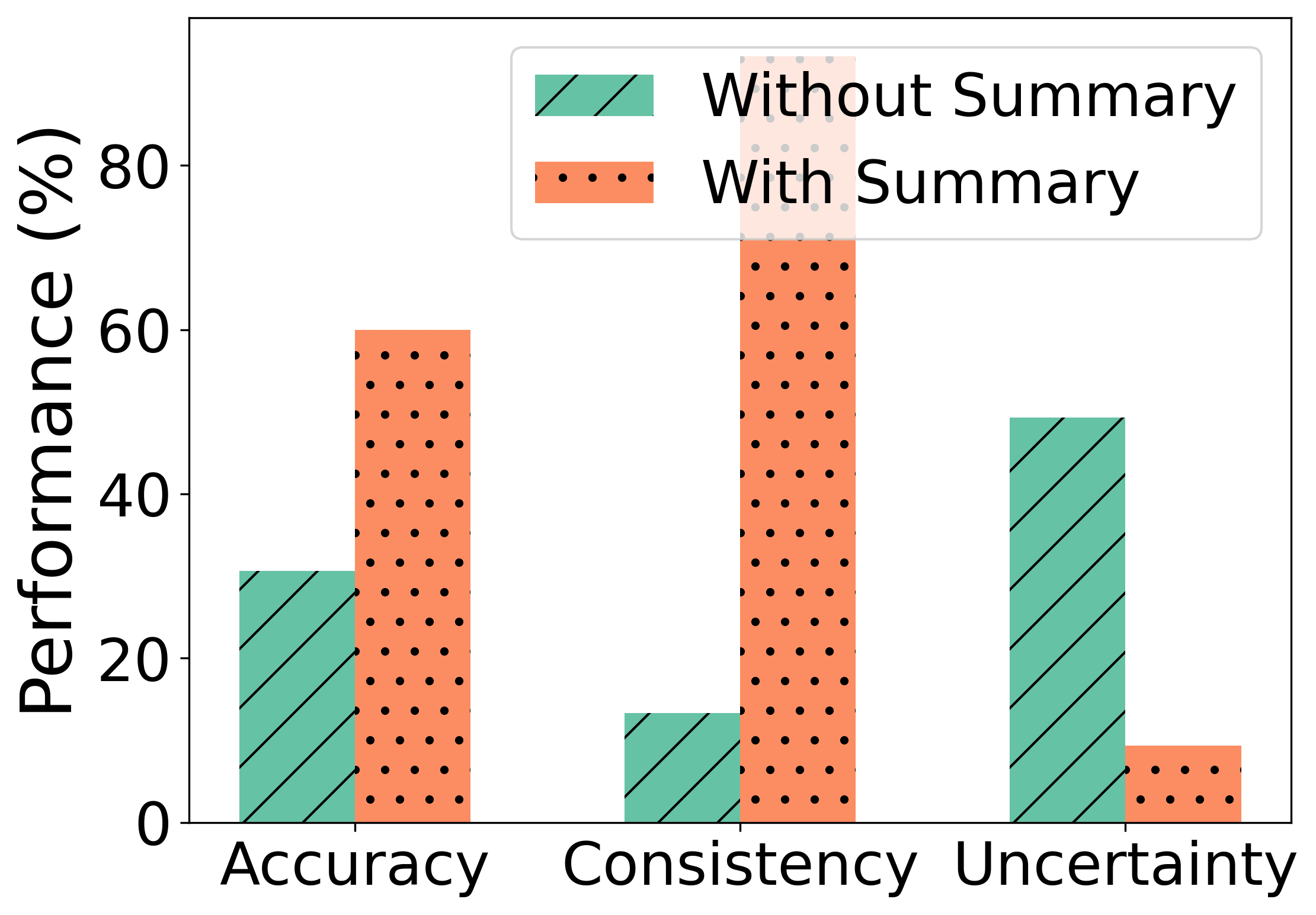}
     \caption{Performance with or without summarization.}
     \label{fig:accuracy_summary}
    \end{minipage}\hspace{1pt}
  \begin{minipage}{.52\linewidth}
    \setlength{\abovecaptionskip}{0.cm}
    \setlength{\belowcaptionskip}{-0.cm}
    \centering
    \includegraphics[width = \textwidth]{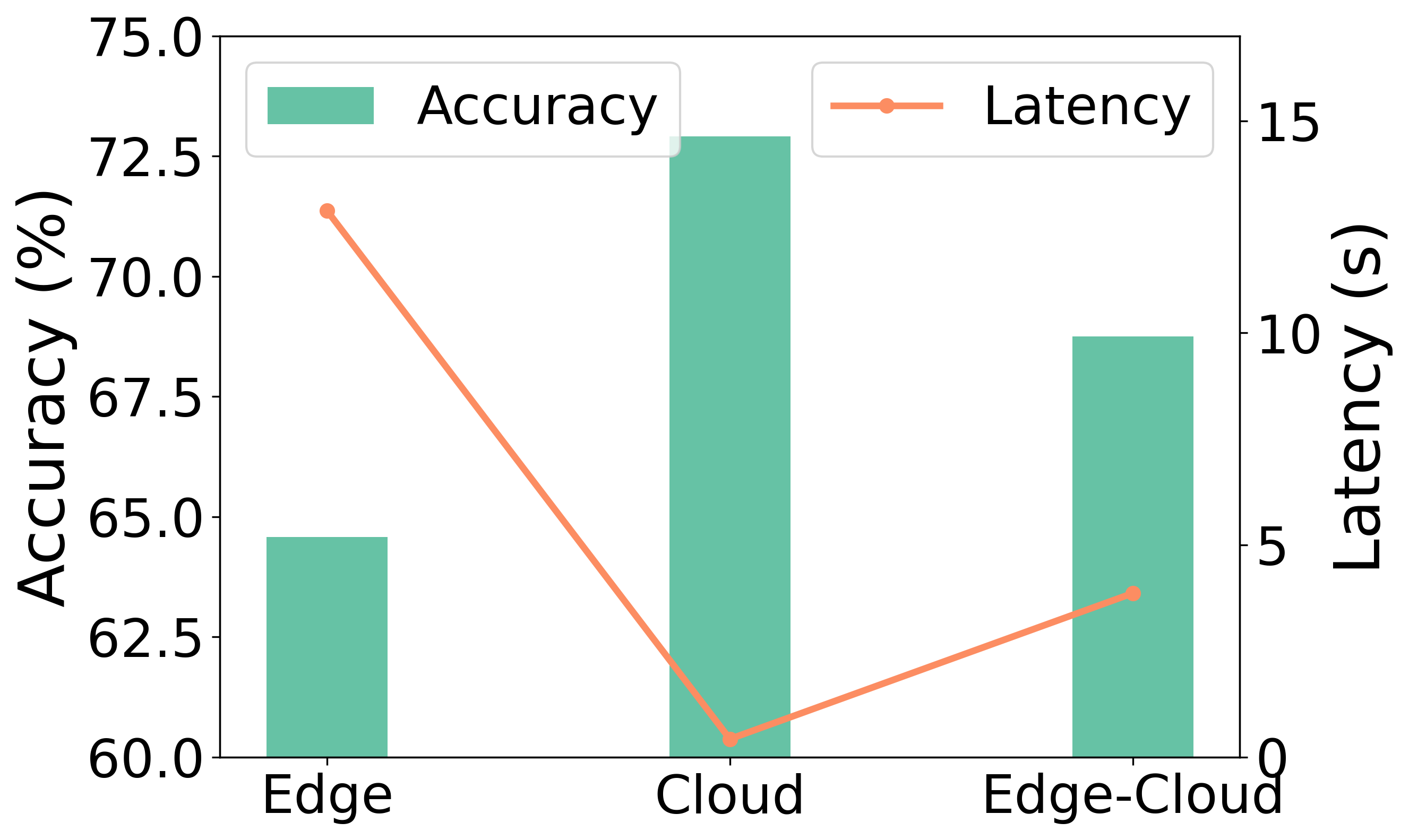}
    \caption{Latency of running LLMs on edge or cloud.}
    \label{fig:latency}
  \end{minipage}
  \vspace{-10pt}
\end{figure}

\subsubsection{Understanding the Performance of LLMSense} We now evaluate the performance of LLMSense with various configurations to understand the effectiveness of our design. 


\textbf{Performance with or without summarization.} Figure \ref{fig:accuracy_summary} shows the performance with or without summarization on the dementia diagnosis task with the Llama-13B model. Compared with results derived from raw sensor traces (i.e., without summary), predictions based on summarization of traces (i.e., with summary) exhibit superior accuracy and consistency. 
Moreover, when presented with raw sensor traces, LLMs often produce ambiguous predictions due to the extensively long traces. However, employing summarized patterns of sensor data markedly reduces the uncertainty of LLMs' predictions. 
This shows substantial performance enhancement achieved by summarization, facilitating LLMs in making more precise and resilient predictions on lengthy sensor traces.


\textbf{Performance with different lengths of history traces.} Figure \ref{fig:accuracy_summary} shows the performance with different lengths of history traces on the occupancy tracking tasks with the Llama-13B model. Here the input data of LLMs is original traces of the latest 30 minutes and traces in the previous 3, 6, 12 and 24 hours respectively. The results show that adding more historical information helps improve the performance of the task. However, the performance does not always increase with the lengths of historical traces, as they may also introduce more noise. When we selectively add historical information for each sample, the performance will be significantly improved, e.g., achieving 87.5\% accuracy in occupancy tracking. 

\subsubsection{Latency} 
Due to privacy concerns, the raw sensor traces may not be allowed to send to the cloud that runs a larger LLM (e.g., GPT3.5) to make the prediction. 
Figure \ref{fig:latency} compares the accuracy and latency of different paradigms. Here ``edge'' means running both summarization and reasoning with LLama2-13B on a local machine, ``cloud'' means running both summarization and reasoning with GPT3.5 on the server, and ``edge-cloud'' means running summarization with LLama2-13B locally, and reasoning with GPT3.5 on the server. The results show that running a larger LLM on the cloud will incur less latency and a higher accuracy than running a smaller LLM on the edge, while the cloud-based solution requires sharing users' raw sensor traces. The edge-cloud paradigm provides a trade-off between accuracy, latency, and data privacy, by only sending the summarization results to the cloud.



\section{Discussion}

We now discuss several directions as future work.

\textbf{Process long or even infinite traces.} While we've developed two new approaches to enhance LLM performance on over long-term sensor traces, interpreting lengthy, or potentially infinite, traces remains a challenge due to the contextual limits of LLMs. In the future, we will incorporate stateful LLMs \cite{gu2021efficiently,gu2023mamba} for high-level reasoning or adaptively select input traces in an online manner.

\textbf{Improve the performance of LLMs based on verifications.} As shown in our evaluations, the prediction results of LLMs may still be consistent and uncertain, especially for complex tasks that require high-level reasoning abilities. In the future, we will study how to quantify the uncertainty or errors of LLMs' outputs, and iteratively improve the performance of LLMs through verifications based on rules or human feedback.

\textbf{Joint optimization of low-level perception and high-level reasoning tasks.} Conventional neural networks designed for low-level perception results require extensive labeled training data, which may not be available in practical applications. For example, annotating every second of videos or wearable data is very difficult. On the other hand, labels for high-level reasoning tasks tend to be sparse. We could study how to leverage the reasoning ability of LLMs to associate low-level perception and high-level reasoning tasks, and enhance the training of low-level perception models.




\section{Conclusion}

In this paper, we propose to harness Large Language Models (LLMs) to analyze observations derived from long-term sensor traces for high-level reasoning tasks.
We design an effective prompting framework for LLMs on high-level reasoning tasks, which can deal with traces from the raw sensor data as well as the low-level perception results. Moreover, we design two strategies to enhance the performance with long sensor traces, including summarization before reasoning and selective inclusion of historical traces. 
This paper provides a few insights for leveraging LLM for high-level reasoning on sensor traces and highlights several directions for future work.

\section*{Acknowledgment}

The research reported in this paper was sponsored in part by: the Air Force Office of Scientific Research under Cooperative Agreement \#FA95502210193; DARPA ANSR Program under Contract \#FA875023C0519; the DEVCOM ARL under Cooperative Agreement \#W911NF-17-2-0196; and, the NIH mDOT Center under Award \#1P41EB028242. The views and conclusions contained in this document are those of the authors and should not be interpreted as representing the official policies, either expressed or implied, of the funding agencies.

\bibliographystyle{IEEEtran}
\bibliography{reference}

\end{document}